\documentclass[10pt, a4paper]{article}
\usepackage{lrec2022} 
\usepackage{multibib}
\newcites{languageresource}{Language Resources}
\usepackage{graphicx}
\usepackage{tabularx}
\usepackage{soul}
\usepackage{titlesec}
\titleformat{\section}{\normalfont\large\bfseries\center}{\thesection.}{1em}{}
\titleformat{\subsection}{\normalfont\SmallTitleFont\bfseries\raggedright}{\thesubsection.}{1em}{}
\titleformat{\subsubsection}{\normalfont\normalsize\bfseries\raggedright}{\thesubsubsection.}{1em}{}
\renewcommand\thesection{\arabic{section}}
\renewcommand\thesubsection{\thesection.\arabic{subsection}}
\renewcommand\thesubsubsection{\thesubsection.\arabic{subsubsection}}

\usepackage{epstopdf}
\usepackage[utf8]{inputenc}
\usepackage{hyperref}
\usepackage{xstring}
\usepackage{float} 
\usepackage{color}
\usepackage{arabtex}
\usepackage{utf8}
\usepackage{makecell}
\usepackage{multirow}
\usepackage{subcaption}
\usepackage{bm}

\title{Wojood: Nested Arabic Named Entity Corpus and Recognition using BERT}

\name{Mustafa Jarrar, Mohammed Khalilia, Sana Ghanem}

\address{Birzeit University \\
         Palestine \\
         \{mjarrar, mkhalilia, swghanem\}@birzeit.edu\\
         }

\abstract{
This paper presents Wojood, a corpus for Arabic \textit{nested} Named Entity Recognition (NER). Nested entities occur when one entity mention is embedded inside another entity mention. Wojood consists of about 550K Modern Standard Arabic (MSA) and dialect tokens that are manually annotated with 21 entity types including person, organization, location, event and date. More importantly, the corpus is annotated with nested entities instead of the more common flat annotations. The data contains about 75K entities and 22.5\% of which are nested. The inter-annotator evaluation of the corpus demonstrated a strong agreement with Cohen's Kappa of 0.979 and an F1-score of 0.976. To validate our data, we used the corpus to train a nested NER model based on multi-task learning using the pre-trained AraBERT (Arabic BERT). The model achieved an overall micro F1-score of 0.884. Our corpus, the annotation guidelines, the source code and the pre-trained model are publicly available.
 \\ \newline }

\begin{document}
\setcode{utf8}
\maketitleabstract

\section{Introduction}
\begin{figure}[H]
    \centering
    \includegraphics[scale=0.6]{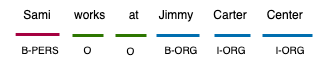}
    \caption{Flat NER Example}
    \label{flat_ner_example}
\end{figure}

Named Entity Recognition (NER) is integral to many Natural Language Processing (NLP) applications such as chatbots and question answering \cite{shaheen2014arabic}, information retrieval \cite{guo2009named}, machine translation \cite{hassan2005integrated}, word-sense disambiguation \cite{HJ21b}, clustering \cite{ARJ17}, interoperability \cite{JDF11}, topic modeling and event discovery \cite{Feng2018}, among others. NER is the task of identifying named entity mentions in unstructured text and classify them to a predefined categories such as \textit{person}, \textit{organization}, \textit{event}, \textit{location}, or \textit{date}. For example, in a sentence like "Sami works at Jimmy Carter Center", "Sami" would be tagged as a \textit{person}, and "Jimmy Carter Center" as an \textit{organization}, as shown in Figure \ref{flat_ner_example}. We use the IOB2 tagging scheme, where \textit{B} indicates the beginning of the entity mention, \textit{I} indicates the inside token, and \textit{O} refers to outside entity \cite{10.3115/977035.977059}. The NER task becomes a sequential labeling task, where each token is tagged with a label based on the contextual information surrounding it \cite{sohrab-miwa-2018-deep}.

Many approaches were proposed for NER in many languages, including Arabic, which we will review in Section \ref{sec:related_work}. However, less attention is given to \textit{nested named entities} \cite{Finkel2009}, which are entity mentions contained inside other entity mentions. For example, "Jimmy Carter" is a \textit{person} that is inside the "Jimmy Carter Center" \textit{organization}, and "Birzeit" is a \textit{geopolitical} entity inside the organization mention "Birzeit University". More examples in Arabic are shown in Figure \ref{fig:nested_ner_example}. 

There are a few reasons why nested NER is understudied. First and most important is lack of data, specially for low-resource languages like Arabic \cite{DH21,J21}. In English, there are a few datasets tagged with nested entities. The GENIA molecular biology corpus \cite{Ohta2002} with 400K tokens and almost 100K annotations, but only 17\% of the entities are embedded in another entity \cite{Katiyar2018}. The ACE2005 which contains about 30K mentions among which there are 3,300 nested mentions \cite{walker2006ace,lu-roth-2015-joint}. The NNE newswire corpus with 279K mentions, 114 entity types with up to six layers of nesting \cite{Ringland2020}. However, in Arabic language, which is the focus of this paper, there are a handful of datasets annotated with flat entities, but none are annotated with nested entities. Second, nested NER adds complexity to the machine learning models in terms of computation, run-time and accuracy \cite{Finkel2009,sohrab-miwa-2018-deep}. Such models need to recursively predict the next tag. Third, data annotation is very expensive and time consuming. An effective nested NER dataset should cover multiple domains, be large in size and have multiple layers of nesting. Also, nested entity annotations are more prone to human errors than flat annotations, since there are multiple layers, which means we need to invest more in inter-annotator agreement to ensure high-quality data.

The goal of this paper is to provide some solutions to overcome some of the challenges we mentioned earlier. Our main contribution is Wojood\footnote{\scriptsize{ Wojood Corpus:  \url{https://ontology.birzeit.edu/wojood}}}, a relatively large-scale Arabic nested named entity corpus. The corpus consists of 550K tokens, which were manually annotated. The corpus supports 21 entity types, four layers of nesting and covers both MSA and dialect in multiple domains (see statistics in Table \ref{table:raw_corpus_stat}). Our evaluation of the inter-annotator agreement shows an overall $Kappa$ of  0.979 and an F1-score of 0.976. Furthermore, the paper presents a method for fine-tuning AraBERT (Arabic BERT) on a multi-task learning architecture to predict nested tags. The model achieved an overall micro F1-score of 0.884.

The rest of this article is organized as follows: Section \ref{sec:related_work} presents related work, Section \ref{sec:corpus} presents our proposed corpus, Section \ref{sec:model} 
describes the BERT-based multi-task architecture and results, Section \ref{sec:implementation} gives overview of the implementation of our NER web service, and we conclude in Section \ref{sec:conclusion}.

\section{Related Work}
\label{sec:related_work}
There are a few available Arabic NER corpora (summarized in Table \ref{table:Related_NER_corpora}). However, none of them supports nested entities and most of them are small in size and support only a small number of entity types. The richest corpus is Ontonotes 5, which is available through the LDC catalog with restricted access \cite{ontonotes5}. It consists of $\sim$300K tokens, collected from MSA media sources and annotated with 18 entity types. The ANERCorp corpus was also collected from MSA media sources \cite{Benajiba2007}. It contains $\sim$150K tokens and supports only four entity types (person, organization, location, and miscellaneous). The Classical Arabic Named Entity Recognition Corpus (CANERCorpus) is a Classical Arabic of Hadith documents ($\sim$258K tokens), which was annotated manually using 14 entity types; however, most entity types are religion-specific \cite{Salah2018}. Although the corpus contains about 72K entity mentions, about 93\% of these entities are tagged as Allah, Prophet, persons, and numbers. The AQMAR corpus was collected from Wikipedia and covers four domains: history, technology, science, and sports \cite{Mohit2012}. The corpus consists of $\sim$47K tokens and was annotated with open-ended types of entities. Furthermore, \newcite{Mostefa2009} presented a multilingual NER corpus that was collected from news articles and included 56 million Arabic tokens. The corpus was, first, automatically annotated with five entity types using a rule-based approach, then a subset of the annotations was verified manually. 

Compared with these corpora, Wojood is larger, supports nested entities, 21 entity types, and multiple domains (health, finance, technology, law, elections, politics, migration, terrorism, history, culture, and social media topics). Our corpus also includes MSA and dialectal text.

\begin{table}[ht]

\begin{center}
\scriptsize
\begin{tabular}{|m{0.082\textwidth}|m{0.025\textwidth}m{0.025\textwidth}  m{0.025\textwidth}m{0.030\textwidth}m{0.043\textwidth}m{0.063\textwidth}|}\hline 
\textbf{Corpus}&\textbf{Nested}&\textbf{Tokens}&\textbf{Entities}&\makecell[c]{\textbf{Classes}}&\textbf{Arabic}&\textbf{Domain}\\
\hline \hline 
Ontonotes5 & No & 300k & 28k & 18 & MSA & News \\     \hline
ANERCorp  & No & 150k & 11k & 4 & MSA & News \\     \hline
Canercorpus  & No & 258k & 72k & 14  & Classic & Religion \\     \hline
AQMAR & No & 74k & - & open & MSA & 4 domains\\     \hline \hline
\textbf{Wojood Corpus}  & \textbf{YES} & \textbf{550K} & \textbf{75K} & \textbf{21} & \textbf{MSA-Dialect} & \textbf{Multi} \\     \hline

\end{tabular}
\caption{Existing Arabic NER Corpora}
\label{table:Related_NER_corpora}
 \end{center}
\end{table}

For entity recognition, several NER approaches have been proposed for Arabic as reviewed in \cite{Shaalan2014,Ali2020}. NER approaches in Arabic can be grouped into rule-based, machine learning, and advanced deep learning approaches. Rule-based approaches use linguistic features (context, morphology, character-level, etc.)
and maybe gazetteers for detecting entities \cite{shaalan2007,Jaber2018}. Machine Learning methods such as support vector machines (SVM) and conditional random field (CRF) were also used for NER with less to no need for language-specific features \cite{Benajiba2007,Abdulhamid10,Mohit2012,abdallah2012}. More recently, Deep Learning approaches have shown better results in recognizing entities. For example, an F1-score of 0.8877 was achieved with the ANERcorp corpus using convolution neural networks (CNN) and bi-directional long-short term memory (BiLSTM) \cite{Khalifa2019}. Experiments with BERT models, using the same ANERcorp corpus, achieved 0.8438 and 0.8913 F1-score with AraBERT2 \cite{Antoun2020} and ARBERT \cite{ARBERT}, respectively. A CRF layer was added to AraBERT2 by \cite{Qurishi21}, which increased the F1-Score to 0.912. Nevertheless, none of these approaches tackled the problem of recognizing nested Arabic named entities due to the absence of such corpora for Arabic.

NER has achieved a remarkable accuracy on many datasets in high-resource languages such as English. Models were published on CoNLL-2002 \cite{tjong-kim-sang-2002-introduction}, CoNLL-2003 \cite{tjong-kim-sang-de-meulder-2003-introduction}, Ontonotes \cite{weischedel2011ontonotes} and i2b2 clinical dataset \cite{uzuner20112010}. CoNLL has approximately 320K tokens annotated with four entity types: person, location, organization and miscellaneous \cite{passos-etal-2014-lexicon}. Ontonotes contains about 1.6M tokens annotated with 18 entity types \cite{passos-etal-2014-lexicon} and 2010 i2b2 dataset contains about 416K tokens \cite{dai-adel-2020-analysis}.

One of the most common model architectures for NER models is based on LSTM and CRF \cite{lample-etal-2016-neural}. LSTM based architecture was extended by others for clinical joint entity extraction and assertion detection such as negation  \cite{bhatia-etal-2019-joint}. Biomedical NER based on Recurrent Neural Network (RNN) was proposed in \cite{7359761}. LSTM and CNN based models rely on pre-trained word embeddings such as Word2Vec \cite{word2vec}, GloVe \cite{pennington-etal-2014-glove} and FastText \cite{fasttext}.

More recently, \cite{devlin-etal-2019-bert} proposed a new language representation model, bi-directional representation from transformers (BERT). Transformer based language models (LM) rely on attention to capture the contextual information and they can model longer dependencies. A common BERT model for NER usually consists of the BERT encoder to encode each token in the input sequence into a 768 dimensional vectors, which are used as input to a linear layer, followed by an activation function such as softmax to output a probability distribution of the tags for each token.

Although nested NER is more complex, it started to receive greater attention in recent years in languages like English, thanks for the advancements in machine learning and deep learning. A multi-layer BiLSTM that learns hypergraph representation for nested entities proposed in \cite{Katiyar2018}, which was trained on ACE2004 and ACE2005 datasets and the authors reported 0.727 and 0.705 F1-score, respectively.  A stacked flat LSTM model proposed by \cite{ju-etal-2018-neural} for nested tag prediction. The model consists of an embedding layer, followed by stacked flat NER units, each unit contains an LSTM and CRF layer. The number of stacked layers is dynamically added until the model predicts "O" tags for all tokens in the sequence. The output of the current flat NER layer are merged to form a representation of the detected entities, which are fed into the next flat NER layer. The model assumes the sequence is tagged once with the same entity type. The author reported F1-score of 0.771 on the nested GENIA molecular biology corpus \cite{Ohta2002}. A similar approach to \cite{ju-etal-2018-neural} was proposed by \cite{wang-etal-2020-pyramid} uses a pyramid architecture. The sequence is recursively encoded through stacked flat NER layers from bottom to top. Similarly, the model relied on LSTM architecture to encode the sequence. Their model was trained on ACE2004, ACE2005, GENIA, and NNE newswire, with 0.8774, 0.8634, 0.7931, and 0.9468 F1-scores, respectively.

Another LSTM based model published by \cite{sohrab-miwa-2018-deep} uses an exhaustive approach. The model exhaustively considers all possible regions in the sequence, however, the maximum number of regions is user pre-defined. Their approach allows for predicting nested tag sequences for entities of the same type. The authors in \cite{10.1162/tacl_a_00334} proposed a nested NER LSTM-CRF model that uses second-best sequence strategy to predict nested tags. In their approach, the LSTM layer acts as the encoder, followed by CRF for sequence decoding. Their approach uses the probabilities returned by CRF to compute the next best sequence of tags. The list above is certainly not exhaustive, but it illustrated the need for nested NER datasets and models in the Arabic language.

\section{The Corpus}
\label{sec:corpus}
\subsection{Corpus Collection and Preparation}


Our corpus was collected from multiple sources, various domains and covers a number of different topics (Table \ref{table:raw_corpus_stat}). Subset of the corpus ($\sim$258K tokens) was manually collected from online articles published on websites including \href{https://www.un.org/}{un.org}, \href{https://www.hrw.org/}{hrw.org}, \href{https://msf.org/}{msf.org}, \href{https://www.who.int/}{who.org}, \href{https://mipa.institute/ar}{mipa.institute}, \href{https://www.elections.ps/}{elections.ps}, \href{https://sa.usembassy.gov/}{sa.usembassy.gov}, \href{https://www.diplomatie.ma/ar}{diplomatie.ma}, and \href{https://www.quora.com/}{quora.com}. This subset covers several topics including health, information and communication technologies (ICT), finance, elections, law, politics, migration, and terrorism. The second subset ($\sim$227K tokens) was extracted from Awraq, the Birzeit University digital Palestinian archive, which covers modern history and cultural heritage. We extracted the descriptions of about 17K documents. Each description is one sentence (4-30 tokens) describing the content of a document. The importance of this subset is that it covers many types of entities and about 60\% of the tokens are entities, most of which are overlapping. The third subset ($\sim$65K tokens) is based on the Palestinian \cite{JHRAZ17,JHAZ14} and Lebanese \cite{EJHZ22} corpora, a dialect text collected from social media and other Levant resources discussing general topics. We aimed at generating a corpus that contains both dialect text and MSA for diversification.

No cleaning, normalization, or pre-processing steps were performed. The raw text was segmented into sentences, each is given a sentence ID. Each sentence was then tokenized, and each token is assigned an index to represent its position in the sentence. The corpus was then saved using a comma separated values (CSV) format and uploaded to Google Sheets for annotation. The Sheet is a table of four columns: sentence ID, token position, token, and the NER tags. The annotators were asked to read the sentence vertically, and tag each token in the fourth column. 
Table \ref{table:nested_annotation_guideline} shows a tokenized sentence and its tags.

\begin{table}[ht]
\begin{center}
\footnotesize
\begin{tabular}{| m{4cm}  |r r| }
      \hline 
      \textbf{Source} - \textbf{Topics} & \textbf{Sentences} & \textbf{Tokens} \\
      \hline \hline
     \makecell[l]{Web Articles  (MSA) \\ {\scriptsize Health, Finance, ICT, Law, Elections,} \\ {\scriptsize  Politics,  Migration and Terrorism}}
          & 9,053 & 258,102\\       \hline
      \makecell[l]{Archive (MSA) \\ {\scriptsize  History and Culture}} & 12,271  & 227,020 \\       \hline
      \makecell[l]{Social Media (Dialect)  \\ {\scriptsize General topics}} & 5,653  & 65,342 \\       \hline \hline 
     \multicolumn{1}{|r|}{\textbf{Total}} & \textbf{26,977} & \textbf{550,464}\\       \hline
\end{tabular}
\caption{Statistics about the raw corpus}
\label{table:raw_corpus_stat}
 \end{center}
\end{table}

\begin{figure}
\centering
\begin{subfigure}[b]{0.55\textwidth}
   \includegraphics[scale=0.6]{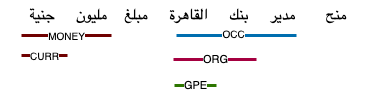}
   \caption{}
   \label{fig:nested_ner_example_a} 
\end{subfigure}

\begin{subfigure}[b]{0.55\textwidth}
   \includegraphics[scale=0.54]{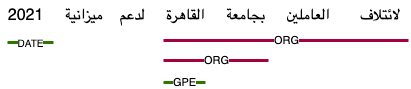}
   \caption{}
   \label{fig:nested_ner_example_b}
\end{subfigure}

\caption[Two numerical solutions]{Example of nested annotations. (a) nested entity of mentions of different types ("Bank of Cairo Manager was awarded one million pound"), (b) illustrates nested entity mentions with same type ("Employees' union at Cairo University to support 2021 budget").}
\label{fig:nested_ner_example}
\end{figure}



\subsection{Annotation Process}
Twelve annotators were carefully selected to participate in the annotation process. All annotators are 4\textsuperscript{th}-year students with high distinction at Birzeit University. A half-day workshop was organized to train about 30 candidate annotators and explain the annotation guidelines; then, each was given a quiz of 750 tokens to annotate, which we designed to be hard. The 12 students who achieved above 90\% in the quiz were recruited. The annotation process, managed by two NER experts, was conducted in three phases over eight months: 

\begin{description}
\item[Phase I:] Each annotator was given about 46K tokens to annotate. The two experts were also helping on critical cases and providing continuous feedback to the annotators during this phase.

\item[Phase II:] The experts reviewed all annotations in phase I and asked the annotators to re-visit some annotations. A program was developed to automatically extracts all entities from the annotated corpus in Phase I. The experts then carefully reviewed all extracted entities. In case of errors, doubts, or incomplete annotations, feedback is provided by the experts to the annotators. The goal here is to improve the consistency of the annotations across all annotators. In other words, our main goal in this phase is not to detect missing annotations; but rather, to validate the entities that have been annotated and increase the alignment among the annotators.

\item[Phase III:] We fine-tuned BERT on a multi-task learning architecture to predict nested entities based on the annotated corpus then used this model to re-annotate the corpus. We compared the manual annotations with the annotations suggested by the model and asked the annotators to verify the mismatches. We repeated this step twice. With this strategy we were able to detect some missing annotations.
\end{description}

Table \ref{table:tag_counts} provides some statistics about the final annotations and the frequency of each entity type. The final dataset contains about 75K entities, out of which 17K entities (22.5\%) are nested within other entity mentions and only 576 of the nested entities are of the same type. Among all entity types, ORG, OCC and FAC are the entities that overlap the most.


\begin{table}[ht]
\footnotesize

\begin{center}
\begin{tabular}{| wl{1.8cm}| wr{1.3cm} wr{1.3cm}| wr{1cm}| }
\hline 
\makecell[c]{\textbf{Tag}} & \makecell[c]{\textbf{Count {\tiny Flat}}} & \makecell[c]{\textbf{Count {\tiny Nested}}}&  \makecell[c]{\textbf{Total}}\\ \hline \hline
PERS &6531 & 739 & 7,270 \\ \hline
NORP &4,928 & 334 & 5,262 \\ \hline
OCC &5,351 & 164 & 5,515 \\ \hline
ORG &15,292 & 3,493 & 18,785 \\ \hline
GPE &11,501 & 10,279 & 21,780 \\ \hline
LOC &755 & 162 & 917 \\ \hline
FAC &939 & 276 & 1,215 \\ \hline
PRODUCT &54 & 1 & 55 \\ \hline
EVENT &2649 & 123 & 2,772 \\ \hline
DATE &2,398 & 105 & 2,503 \\ \hline
TIME &331 & 2 & 333 \\ \hline
LANGUAGE &197 & 1 & 198 \\ \hline
WEBSITE &607 & 0 & 607 \\ \hline
LAW &496 & 0 & 496 \\ \hline
CARDINAL &1,790 & 23 & 1,813 \\ \hline
ORDINAL &4,041 & 989 & 5,030 \\ \hline
PERCENT &137 & 0 & 137 \\ \hline
QUANTITY &49 & 8 & 57 \\ \hline
UNIT &5 & 54 & 59 \\ \hline
MONEY &197 & 30 & 227 \\ \hline
CURR &25 & 216 & 241 \\ \hline \hline
\makecell[r]{\textbf{Total}} & \textbf{58,273} & \textbf{16,999} & \textbf{75,272} \\ \hline

\end{tabular}
\caption{Counts of the flat, nested, and total of each entity type in the corpus.}
\label{table:tag_counts}
 \end{center}
\end{table}

\subsection{Annotation Guidelines}
The data was annotated with 21 entity types, see Table \ref{table:tags_desc} for a short description and examples for each entity type. The full documentation of the guidelines is available online\footnote{\scriptsize{our guidelines:  \url{https://ontology.birzeit.edu/wojood}}}. Our annotation guidelines are designed to be compatible with the guidelines proposed by the LDC's OntoNotes 5 guidelines \cite{ontonotes5}. However, we introduced four new tags (occupation, website, unit, currency), and revised the LDC guidelines to be more suitable for Arabic names. For example, we consider prefixes and suffices to be part of entity names, such as ({\scriptsize \<بمحمد> - \<لخان الخليلي> - \<سادسهم> }).

\begin{table}
\small

\begin{center}
\begin{tabular}{| p{1.5cm} | p{5.6cm}| }
\hline
\textbf{Tag} & \textbf{Short Description}  \\ \hline

\multirow{2}{*}{PERS} & People names, including first, middle, last and nicknames. Titles are not included except Prophets, kings, etc. \\ 
& \makecell[r]{{\tiny \<فيروز ، عادل إمام ، ام كلثوم ، ابن احمد ، الملك عبدالله ، النبي محمد> }} \\ \hline

\multirow{2}{*}{NORP} & Group of people. \\ 
& \makecell[r]{{\tiny \<العرب ، المسحيين ، سكان القدس ، مصابي كورونا ، وزراء الخارجية العرب> }} \\ \hline

\multirow{2}{*}{OCC} & Occupation or professional title. \\ 
& \makecell[r]{{\tiny \<الدكتور ، المهندس ، رئيس جامعة الزيتونة، مدير بنك القاهرة ، قائد الجيش> }} \\ \hline

\multirow{2}{*}{ORG} & Legal or social bodies like institutions, companies, agencies, teams, parties, armies, and governments.  \\ 
& \makecell[r]{{\tiny \<بنك القاهر ، شركة أرمكس ، ريال مدريد ، داعش ، الجيش المصري ، حكومة العراق> }} \\ \hline

\multirow{2}{*}{GPE} & Geopolitical like countries, cities, and states.  \\ 
& \makecell[r]{{\tiny \<ليبا ، صفاكس ، مدينة القدس ، الجمهورية اللبنانية ، الولايات المتحدة الأمريكية> }} \\ \hline

\multirow{2}{*}{LOC} & Geographical locations (Non-GPE), rivers, seas, mountains, and geographical regions. \\ 
& \makecell[r]{{\tiny \<البحر الميت ، قناة السويس ، الخليج العربي ، آسيا ، الوطن العربي ، شرق آسيا> }} \\ \hline

\multirow{2}{*}{FAC} & Name of a specific place, like roads, cafes, buildings, airports, and gates .  \\ 
& \makecell[r]{{\tiny \<مطار صنعاء ، سجن ابو غريب ، المسجد الأقصى ، تمثال الحرية ، مقهى الفيشاوي> }} \\ \hline

\multirow{2}{*}{PRODUCT} & Vehicles, weapons, foods, etc.  \\ 
& \makecell[r]{{\tiny \<مرسيدس سي٠٨١ ، ايفون ٣١ ، إم٦١ ، دبابة مركابا ، تروفين ، ونستون> }} \\ \hline

\multirow{2}{*}{EVENT} & Name of an event of general interest, like battles, wars, sports events, demonstrations, disasters, elections,  and national or religious days. The place and date of the event are included as part of the event name.\\ 
& \makecell[r]{{\tiny \<حرب عام 1973 ، القمة العربية لعام 2005 ، عيد الفطر ، يوم الأرض> }} \\ \hline
 
\multirow{2}{*}{DATE} & Reference to specific or relative dates including age, era, duration, month and day names. Charterers used to separate date components are part of the date. \\ 
& \makecell[r]{{\tiny \<2022 ، 13 يونيو ، الاسبوع الماضي ، من 2019 حتى 2020 , الفترة العثمانية> }} \\ \hline

\multirow{2}{*}{TIME} & Reference to specific or relatives times, which is less than a day; including day times like evening and sunset.\\ 
& \makecell[r]{{\tiny \<الساعة ١٢ ، من الساعة الخامسة حتى السابعة مساء ، خلال ساعتين ، صباح الجمعة> }} \\ \hline

\multirow{2}{*}{LANGUAGE} & Named human language or named dialect. \\  
& \makecell[r]{{\tiny \<اللغة العربية ، الفصحى ، الدارجة المغربية ، الإنجليزية ، اللغة الفرنسية> }} \\ \hline

\multirow{2}{*}{WEBSITE} & Any named website or specific URL. \\ 
& \makecell[r]{{\tiny www.schema.org , \<صفحة جامعة الزيتونة ، موقع فيسبوك ، يوتيوب>}} \\ \hline

\multirow{2}{*}{LAW} & Reference to a specific law or part of it.\\ 
& \makecell[r]{{\tiny \<المادة (114) من قانون العقوبات لسنة 2005 ، قانون الاستثمار، الدستور الأردني> }} \\ \hline

\multirow{2}{*}{CARDINAL} & Numerals written in digits or words. \\ 
& \makecell[r]{{\tiny \<150 ، 30 ، 1.5 ، صفر، اثنان ، أربعة وعشرون ، مليون> }} \\ \hline

\multirow{2}{*}{ORDINAL} & Any ordinal number, in digits or words, that does not refer to a quantity. \\ 
& \makecell[r]{{\tiny A2341 , \<125 ، الخامس ، الأول ، ثانيا، ثالثا ، الرابع والعشرين> }} \\ \hline

\multirow{2}{*}{PERCENT} & A word or a symbol referring to a percent.  \\ 
& \makecell[r]{{\tiny \<5 بالمئة ، 10\% ، 9 من كل الف> }} \\ \hline

\multirow{2}{*}{QUANTITY} & Any value measured by standardized units, except dates, times, and money.\\  
& \makecell[r]{{\tiny \<٣ كيلومتر، مئة قدم ، 3 طن ، 50 غرام ، 25 سم مربع> }} \\ \hline 
\multirow{2}{*}{UNIT} & Any name or symbol of a standardized unit.\\ 
& \makecell[r]{{\tiny \<ميل ، كيلو ، كيلومتر ، إنش ، كيلوغرام ، هكتار ، مل ، بيكو> }} \\ \hline

\multirow{2}{*}{MONEY} & Absolute monetary quantity, including currency names. \\ 
& \makecell[r]{{\tiny 6€ , \<مئة وخمسون درهم اماراتي ، اثنان وثلاثون يورو ، 8 دولار >}} \\ \hline

\multirow{2}{*}{CURR} & Any name or symbol referring to currency.\\ 
& \makecell[r]{{\tiny \$ , € , \<دولار ، جنيه مصري ، دينار ، فرنك ، ريال سعودي >}} \\ \hline

\end{tabular}\caption{NER classes}
\label{table:tags_desc}
 \end{center}
\end{table}

\textbf{Nested annotations} are the core of our corpus. If an entity mention is part of another entity mention, both are annotated. Figure \ref{fig:nested_ner_example} illustrates examples of nested annotations. For tokens that are annotated with multiple labels, these labels are sorted from outside to inside or top to bottom. For example, in Figure \ref{fig:nested_ner_example_a} the token (\textit{{\scriptsize \<بنك> \slash bank}}) is labeled first with {\scriptsize I-OCC } then {\scriptsize B-ORG} because the {\scriptsize OCC} is the top-most entity; thus, ({\scriptsize B-ORG I-OCC}) would be incorrectly ordered.

In some cases, which are rare, two entities of the same type may overlap, as illustrated in Figure \ref{fig:nested_ner_example_b}, which shows that
(\textit{{\scriptsize \<ائتلاف العاملين بجامعة القاهرة>  \slash The employee union of Cairo University}})
  is an organization that overlaps with  
 (\textit{{\scriptsize \<جامعة القاهرة> \slash Cairo University}})
 which is another organization. 
When annotating nested entities of the same type, some token may have duplicate tags. Table \ref{table:nested_annotation_guideline} shows a sentence with three layers of nesting. Notice that (\textit{{\scriptsize \<القاهرة> \slash Cairo}}) is part of three mentions, two of which are {\scriptsize ORG}. (\textit{{\scriptsize \<القاهرة> \slash Cairo}}), which happened to be the last token in the entity mention is thus assigned {\scriptsize I-ORG} in both the 1\textsuperscript{st} and 2\textsuperscript{nd} layers. 

\begin{table}
\small
\center
\begin{tabular}{|l|l|l|l|}
      \hline 
      \textbf{Token} & \textbf{1\textsuperscript{st} Layer} & \textbf{2\textsuperscript{nd} Layer} & \textbf{3\textsuperscript{rd} Layer}  \\
      \hline 
      \textit{\<لائتلاف>  \slash Union of} & B-ORG & & \\
      \textit{\<العاملين> \slash  employees} & I-ORG & & \\
      \textit{\<بجامعة> \slash at University} & I-ORG & B-ORG & \\
      \textit{\<القاهرة> \slash of Cairo} & I-ORG & I-ORG & B-GPE  \\
      \textit{\<لدعم> \slash to support} & O & & \\
      \textit{\<ميزانية> \slash budget} & O & &  \\
      2021 & B-DATE & & \\
      \hline 
\end{tabular}
\caption{Examples of nested entity annotations}
\label{table:nested_annotation_guideline}
\end{table}

\subsection{Inter-annotator Agreement}

To evaluate the Inter-Annotator Agreement (IAA), we asked each annotator to annotate 2,000 tokens that were previously annotated by a different annotator, and which we selected randomly, from different sources and domains. We also paid attention to select sentences from all domains. In this way, the IAA is performed between: $A_{1}-A_{2}, A_{2}-A_{3},..,A_{12}-A_{1}$, with a total of about 24K token (4.3\% of the corpus). Table \ref{table:IAA} presents IAA for each entity type.

Although IAA is typically measured using Cohen’s $Kappa$, some researchers did not recommend it for evaluating named entity annotations \cite{Hripcsak2005,Campillos2021}. They noted that named entities are sequences of tokens and do not contain negative cases. As stated in \cite{Hripcsak2005}, "\textit{$Kappa$ statistics and reliability coefficients cannot be calculated in studies without a negative case count}". Additionally, since the corpus might be unbalanced, as the number of tokens tagged "O" is much higher than other labels, $Kappa$ would be high and overestimates IAA. On the other hand, ignoring the "O" label yields low $Kappa$ scores \cite{Hripcsak2005,Brandsen2020,Campillos2021,Claudia2022}. Instead, they recommended to use the F1-score, as it might be more appropriate in reflecting the IAA in NER tasks. Nevertheless, in this paper, we present the three evaluations: (\textit{i}) $Kappa$ with "O", denotaed as $\kappa_{O}$, (\textit{ii}) $Kappa$ without "O", denoated as $\kappa_{\sim O}$, and (\textit{iii}) F1-score. In fact, we found that the three evaluation scores are close to each other.

\subsection*{Calculating $\kappa_{O}$}
To calculate the $\kappa_{O}$ for a certain tag (e.g., ORG), we count the number of agreements and disagreements between annotators on this tag. All other tags are ignored and counted as "O". Agreements are counted as pair-wise matches at the token level; thus, if a token is annotated, e.g., as ORG by one annotator and as O by another annotator, we count it as a disagreement. As such, $Kappa$ is calculated as the following \cite{Eugenio04}: 
\[ \kappa = \frac{P_o - P_e}{1-P_e} \]
where $P_o$ is the observed agreement between annotators, and $P_e$ is the expected agreement (agreement by chance) defined as the agreement between annotators obtained if they randomly assign tags while annotating. $P_e$ is calculated as:
\[ P_e = \frac{1}{N^2}\sum_{T} {n_{T1}}\times  {n_{T2}}\]
where $n_{Ti}$ is the number of tokens to which annotator $i$ assigned the tag $T$. $N$ is the total number of annotated tokens. The overall $\kappa_{O}$ is calculated using macro-average of all $Kappa$ scores of the other tags.

\subsection*{Calculating $\kappa_{\sim O}$}
The $\kappa_{\sim O}$ for a given tag is calculated similar to the calculations of the $\kappa_{O}$, however, we did not include the agreement on the "O" label.

\subsection*{Calculating F1-score}
To calculate the F1-score for a specific tag $T$ (e.g., PERS), we counted only the tokens labeled with $T$ by at least one of the annotators. Then, we performed a pair-wise comparison and counted the matches of this tag as true positives $TP$ (i.e. number of tokens labeled PERS by both annotators). Otherwise, if the first annotator disagrees with the second, we count those as false negatives $FN$, and if the second disagrees with the first, we count them as false positives $FP$. In this way the number of disagreements is $FN + FP$. Calculating the F1-score for a certain tag $T$ is then given by:

\[F1-Score =\frac{2 TP }{2 TP + FN + FP} \]

The overall F1-score of all annotations is the micro-average of the F1-scores of all tags, which takes into account the count (i.e., weight) of each tag.\\


\begin{table}[h]

\begin{center}
\scriptsize
\begin{tabular}{|wl{1.5cm}|
r  r  r  >{\raggedleft\arraybackslash}m{0.7cm}  >{\raggedleft\arraybackslash}m{0.8cm} wr{0.65cm}|}\hline 
\textbf{Tag} & 
\textbf{TP} & \textbf{FN} & \textbf{FP} & \makecell[c]{\bm{$\kappa_{O}$}} & 
\makecell[c]{\bm{$\kappa_{\sim O}$}} & \makecell[c]{\textbf{F1-Score}}\\\hline \hline 
PERS & 270 & 2 & 1 & 0.994 & 0.994 & 0.994 \\     \hline
NORP & 659 & 29 & 26 & 0.959 & 0.955 & 0.96 \\     \hline
OCC & 486 & 11 & 2 & 0.987 & 0.986 & 0.987 \\     \hline
ORG & 1,713 & 33 & 30 & 0.981 & 0.974 & 0.982 \\     \hline
GPE & 778 & 7 & 13 & 0.987 & 0.985 & 0.987 \\     \hline
LOC & 135 & 7 & 4 & 0.961 & 0.96 & 0.961 \\     \hline
FAC & 48 & 0 & 3 & 0.97 & 0.969 & 0.97 \\     \hline
PRODUCT & 5 & 0 & 0 & 1 & 1 & 1 \\     \hline
EVENT & 386 & 56 & 14 & 0.915 & 0.91 & 0.917 \\     \hline
DATE & 688 & 28 & 8 & 0.974 & 0.971 & 0.975 \\     \hline
TIME & 63 & 8 & 3 & 0.919 & 0.919 & 0.92 \\     \hline
LANGUAGE & - & - & - & - & - & - \\     \hline
WEBSITE & - & - & - & - & - & - \\     \hline
LAW & 257 & 1 & 0 & 0.998 & 0.998 & 0.998 \\     \hline
CARDINAL & 250 & 3 & 6 & 0.982 & 0.982 & 0.982 \\     \hline
ORDINAL & 277 & 1 & 4 & 0.991 & 0.991 & 0.991 \\     \hline
PERCENT & 43 & 0 & 0 & 1 & 1 & 1 \\     \hline
QUANTITY & 6 & 0 & 0 & 1 & 1 & 1 \\     \hline
UNIT & 3 & 0 & 0 & 1 & 1 & 1 \\     \hline
MONEY & 29 & 0 & 0 & 1 & 1 & 1 \\     \hline
CURR & 14 & 0 & 0 & 1 & 1 & 1 \\     \hline
\hline
\thead {\textbf{Overall}}
& \makecell[r]{\textbf{6,110} \\ {\tiny count}} 
& \makecell[r]{\textbf{114} \\ {\tiny count}} 
& \makecell[r]{\textbf{186} \\ {\tiny count}}  
& \makecell[r]{\textbf{0.98} \\ {\tiny macro}}  
& \makecell[r]{\textbf{0.979} \\ {\tiny macro}}
& \makecell[r]{\textbf{0.976} \\ {\tiny micro}}
\\     \hline
\end{tabular}
\caption{Overall inter-annotator agreement for each entity type. }
\label{table:IAA}
 \end{center}
\end{table}

\begin{table*}[t]
\scriptsize
\begin{center}

\begin{tabular}{|l|llllllllllll|} \hline 

\multirow{2}{*}{\textbf{Tag}} & \multicolumn{12}{c|}{\textbf{F1-score}} \\ 
\cline{2-13}
& \textbf{1-2} & 
\textbf{2-3} & 
\textbf{3-4} & 
\textbf{4-5} & 
\textbf{5-6} & 
\textbf{6-7} & 
\textbf{6-8} & 
\textbf{8-9} & 
\textbf{9-10} & 
\textbf{10-11} & 
\textbf{11-12} &
\textbf{12-1}
\\ \hline \hline
PERS & 1 & 1 & 0.97 & 1 & - & 1 & 1 & 1 & 1 & 0.95 & 1 & - \\ \hline
NORP & 0.92 & 0.97 & 0.92 & 0.93 & 0.96 & 0.94 & 100 & 0.97 & 0.95 & 0.96 & 1 & 1 \\ \hline
OCC & 0.98 & 1 & 0.97 & 0.92 & 0.98 & 1 & 1 & 0.98 & 0.99 & 1 & 1 & 1 \\ \hline
ORG & 0.96 & 0.99 & 0.94 & 0.95 & 0.98 & 0.98 & 1 & 1 & 0.98 & 0.99 & 1 & 1 \\ \hline
GPE & 1 & 1 & 0.98 & 0.98 & 0.98 & 0.96 & 1 & 0.99 & 0.95 & 1 & 1 & 1 \\ \hline
LOC & 1 & 1 & 0.92 & 1 & - & 0.91 & 1 & 1 & 0.98 & 1 & - & - \\ \hline
FAC & 1 & - & 1 & 1 & - & 1 & 1 & 1 & 1 & - & - & - \\ \hline
PRODUCT & - & - & - & - & - & 1 & - & - & - & 1 & - & - \\ \hline
EVENT & - & 1 & 0.90 & 0.87 & 0.92 & 0.91 & 1 & 0.92 & 1 & 1 & - & 1 \\ \hline
DATE & 1 & 1 & 0.97 & 0.97 & 1 & 0.92 & 1 & 0.94 & 0.98 & 0.92 & 1 & 1 \\ \hline
TIME & - & 1 & 1 & 1 & - & - & 1 & 1 & - & - & 1 & - \\ \hline
LANGUAGE & - & - & - & - & - & - & - & - & - & - & - & - \\ \hline
WEBSITE & - & - & - & - & - & - & - & - & - & - & - & - \\ \hline
LAW & - & - & - & - & 1 & - & - & - & - & - & 0.99 & 1 \\ \hline
CARDINAL & 0.98 & 1 & 0.97 & 0.99 & 1 & 1 & 1 & 93 & 1 & 0.96 & 1 & 1 \\ \hline
ORDINAL & 0.92 & 1 & 0.97 & 1 & 1 & 0.95 & 1 & 0.96 & 1 & 1 & 1 & 1 \\ \hline
PERCENT & 1 & 1 & 1 & 1 & 1 & 1 & - & - & 1 & - & - & 1 \\ \hline
QUANTITY & - & - & - & - & - & - & - & 1 & - & 1 & - & - \\ \hline
UNIT & - & - & - & - & - & - & - & 1 & - & 1 & - & - \\ \hline
MONEY & 1 & 1 & - & - & - & - & - & - & - & - & - & 1 \\ \hline
CURR & 1 & 1 & - & - & - & - & - & - & - & - & - & 1 \\ \hline
\hline
\thead 
{\textbf{Overall}} & 
{\textbf{0.97}} & 
{\textbf{1}} & 
{\textbf{0.95}} & 
{\textbf{0.94}} & 
{\textbf{0.98}} & 
{\textbf{0.94}} & 
{\textbf{1}} & 
{\textbf{0.97}} & 
{\textbf{0.98}} & 
{\textbf{0.97}} & 
{\textbf{1}} & 
{\textbf{1}} \\ \hline
\end{tabular}
\caption{The F1-score Inter-Annotator Agreement between each pair of annotators (1-2 refers to F1-score between the first and second annotators)}
\label{table:IAA_annotators}
 \end{center}
\end{table*}

\subsection{Annotation Challenges}

The results in Table \ref{table:IAA} illustrate a very strong agreement on all tags. Closer examination of the disagreement scores shows that EVENT, NORP, TIME, LOC, FAC and DATE have the most disagreements. We found that this is mostly due to entity boundaries. Although the annotation guidelines stated that some tokens, like (\textit{{\scriptsize \<بداية> \slash beginning of}}) should be annotated as part of the date, in e.g., (\textit{{\scriptsize \<بداية ٢٠٢١> \slash beginning of 2021}}), but some annotators excluded them. Similarly, (\textit{{\scriptsize \<العرب المهاجرين> \slash Arab emigrants}}) was annotated as one NORP entity by some annotators, while others annotated ({{\scriptsize \<العرب> \slash Arab}})  and (\textit{{\scriptsize \< المهاجرين> \slash emigrants}}) as two separate NOPR entities. Most of the disagreements on ORG were related to the boundary issue as well. For example, the (\textit{{\scriptsize OECD \<منظمة التنمية والتعاون الاقتصادي> \slash Organisation for Economic Co-operation and Development}}), was annotated as one ORG entity by one annotator, while the ({\scriptsize OECD}) was annotated as a separate ORG entity by another annotator. The (\textit{{\scriptsize \<البنك العربي بعمان> \slash The Arab Bank in Amman}}) was sometimes annotated as an ORG entity, while others excluded (\textit{{\scriptsize \<بعمان> \slash Amman}}) from the entity mention.

The EVENT was problematic. Not only that annotators disagree on the boundaries, but they also disagree on the relevancy of the events. For example, one annotator annotated (\textit{{\scriptsize \<اعترفت الولايات المتحدة بالمملكة المتوكلية اليمنية عام 1946> \slash The USA recognized the Mutawakkilite Kingdom of Yemen
in 1946}}) as an EVENT, while others ignored the event completely. Indeed, some events might be important to annotate in some domains but not others.

We believe that the reason of why the overall inter-annotator agreement in our corpus was high is because we managed to provide continuous feedback to the annotators and revise the annotation guidelines if needed. Additionally, the review we carried out in Phase II was critical to improve the consistency among the annotators. It was also beneficial to compare the annotators and the model annotations, specially in the cases where annotators missed, but the model did not.

\section{Named Entity Recognition (NER)}
\label{sec:model}
In addition to the inter-annotator agreement, we developed a model for nested NER to further validated the data and showcase its accuracy. 

\subsection{Multi-Task Learning for Nested NER}
\begin{figure*}
\centering
  \includegraphics[width=0.8\textwidth]{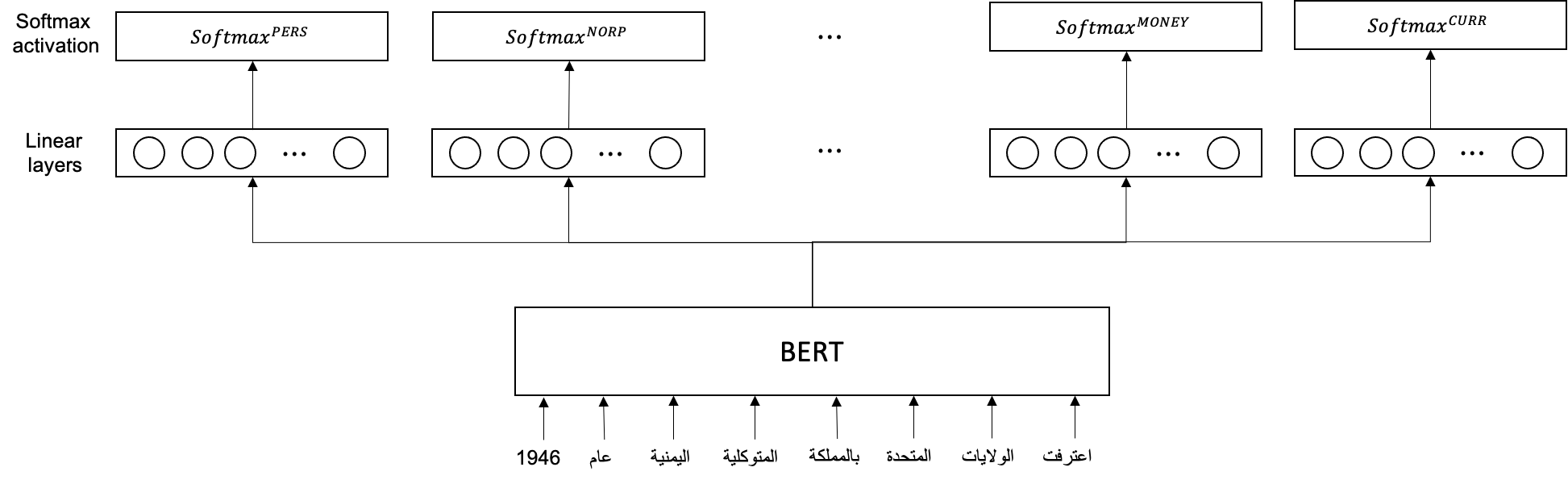}
  \caption{Nested NER model architecture}
  \label{multi_task_ner}
\end{figure*}

This section presents a multi-task based approach for nested NER (Figure \ref{multi_task_ner}). Similar to some of the methods we discussed in section \ref{sec:related_work}, our approach assumes the sequence is tagged once with the same entity type.

The model consists of the sequence encoder and multiple classifiers, one for each entity type. We used transformer encoder based on the BERT language model. There are few BERT models pre-trained on Arabic text including AraBERT \cite{Antoun2020} and ARBERT \cite{ARBERT}. AraBERT was trained on two major datasets, the 1.5 billion words Arabic Corpus \cite{el20161} and the Open Source International Arabic News Corpus (OSIAN), which consists of 3.5 million articles ($\sim1$ billion tokens), from 31 news sources in 24 Arab countries \cite{zeroual-etal-2019-osian}. The final size of AraBERT dataset is 70M sentences, corresponding to about 24GB of text.\\
BERT takes a sequence of tokens as input and for each token outputs a 768 dimensional representation. BERT relies on the attention mechanism \cite{vaswani2017attention} to learn the contextual information of a given token from the entire sequence. The pre-trained BERT used in this paper is the AraBERT. Following the BERT layer are the classification layers, which are linear layers used to project the 768 dimensional token representation into the output space. The multi-task model consists of 21 classification layers, one layer for each entity type listed in Table \ref{table:tags_desc}. Since we use the IOB2 tagging scheme, each linear layer is a multi-class classifier that outputs the probability distribution through softmax activation function for three classes, $C \in \{I, O, B\}$. The model is trained with cross entropy loss objective computed for each linear layer separately, which are summed to compute the final cross entropy loss.

\subsection{Model Training}
The dataset was divided into three different datasets: training (385K tokens, 70\%), validation (55K tokens, 10\%) and test (110K tokens, 20\%). We fine-tuned AraBERT \cite{Antoun2020} on the nested NER tasks with learning rate, $\eta = 1e^{-3}$, batch size of 32, maximum of 20 epochs with early termination if there is no improvement on the validation data after five epochs. In general the model converged around epoch nine. The source code and the nested NER pre-trained model are available on GitHub\footnote{\scriptsize{\url{https://ontology.birzeit.edu/wojood} (PIN=reviewer) the source code will be publicly available pending review decision.}}.

\subsection{Experiments and Results}
We evaluated our model on Wojood, the dataset presented in this paper. However, there is one limitation to the model, it cannot recognize nested entities of same type. For example, in Table \ref{table:nested_annotation_guideline}, the model can tag (\textit{{\scriptsize \<ائتلاف العاملين بجامعة القاهرة>  \slash The Employee Union of the Cairo University}}) with ORG, but cannot tag the sub-entity mention (\textit{{\scriptsize \<بجامعة القاهرة> \slash  Cairo University}}) with ORG. This is a limitation we plan to address in future work.

Overall, the model achieved 0.884 micro F1-score, 0.8772 precision and 0.8909 recall (Table \ref{table:ner_results}). The entity type on which the model was most confident about is GPE with an F1-score of 0.947, while the model was least confident about QUANTITY, UNIT, PERCENT, WEBSITE and TIME with 0.2, 0.25, 0.4426, 0.4936 and 0.5526 F1-score, respectively. The reason for low accuracy on these entity types is the lack of sufficient annotation. As shown in Table \ref{table:tag_counts}, QUANTITY, UNIT and PERCENT have only 57, 59 and 137 mentions, respectively. 

\begin{table}[!ht]
\begin{center}
\footnotesize
\begin{tabular}{|l|ccc|}\hline
\textbf{Tag} & \textbf{Precision} & \textbf{Recall} & \textbf{F1-Score} \\ \hline \hline
PERS & 0.9135 & 0.9122 & 0.9129 \\ \hline
NORP & 0.6828 & 0.7037 & 0.6931 \\ \hline
OCC & 0.7993 & 0.8402 & 0.8193 \\ \hline
ORG & 0.8924 & 0.9072 & 0.8997 \\ \hline
GPE & 0.9424 & 0.9516 & 0.9470 \\ \hline
LOC & 0.8054 & 0.7059 & 0.7524 \\ \hline
FAC & 0.7366 & 0.6481 & 0.6895 \\ \hline
PRODUCT & 0.3333 & 0.2500 & 0.2857 \\ \hline
EVENT & 0.6364 & 0.6488 & 0.6425 \\ \hline
DATE & 0.9253 & 0.9394 & 0.9323 \\ \hline
TIME & 0.6000 & 0.5122 & 0.5526 \\ \hline
LANGUAGE & 0.9310 & 0.7105 & 0.8060 \\ \hline
WEBSITE & 0.4496 & 0.5472 & 0.4936 \\ \hline
LAW & 0.8525 & 0.9123 & 0.8814 \\ \hline
CARDINAL & 0.8437 & 0.8575 & 0.8505 \\ \hline
ORDINAL & 0.9411 & 0.9448 & 0.9430 \\ \hline
PERCENT & 0.2903 & 0.9310 & 0.4426 \\ \hline
QUANTITY & 0.2500 & 0.1667 & 0.2000 \\ \hline
UNIT & 0.5000 & 0.1667 & 0.2500 \\ \hline
MONEY & 0.9143 & 0.8205 & 0.8649 \\ \hline
CURR & 0.8810 & 0.9487 & 0.9136 \\ \hline
\hline

\thead {\textbf{Overall}}
& \makecell[c]{\textbf{0.8772} \\} 
& \makecell[c]{\textbf{0.8909} \\} 
& \makecell[c]{\textbf{0.8840} \\}  
\\     \hline

\end{tabular}
\caption{Nested NER Results}
\label{table:ner_results}
\end{center}
\end{table}

\section{Implementation}
\label{sec:implementation}
\begin{figure}[H]
\centering
 \includegraphics [width=0.48\textwidth] 
 {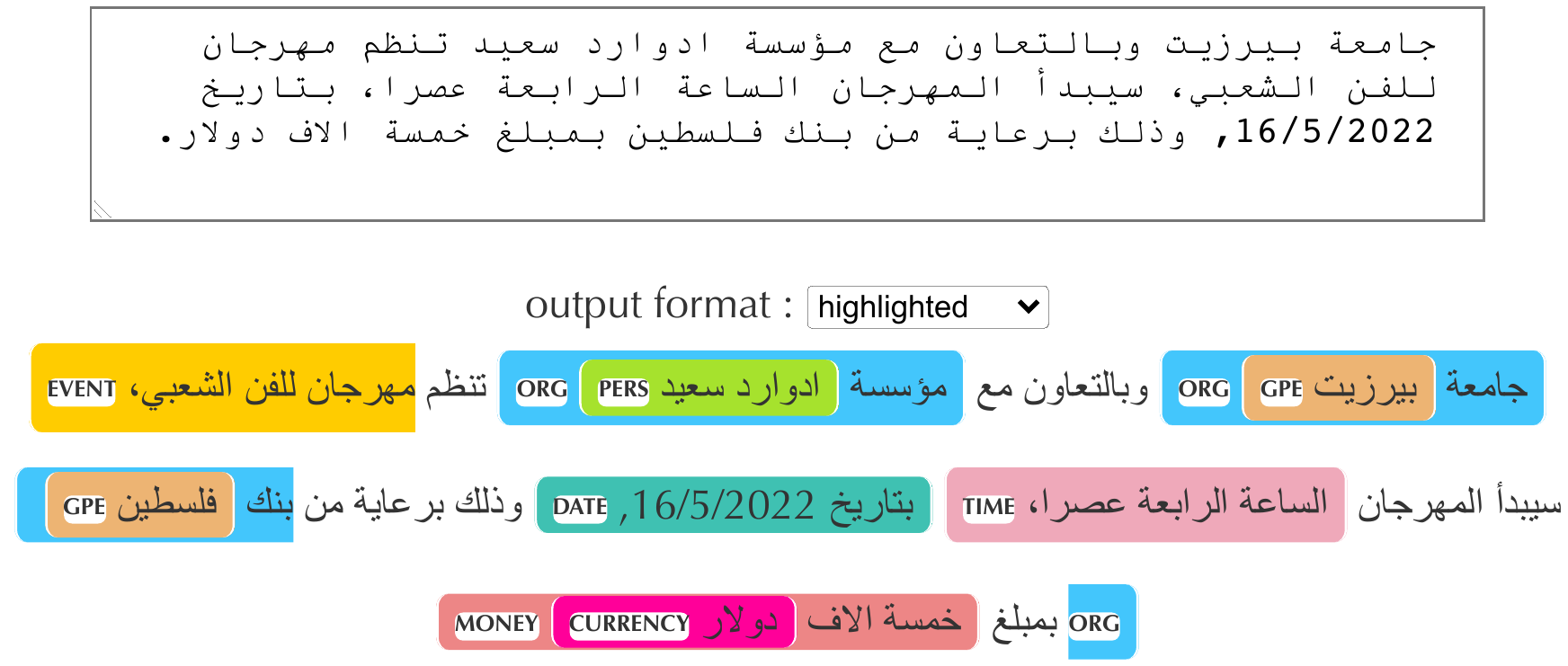}
\caption{Web user interface of the nested named entities.}
\label{fig:demo}
\end{figure}

A RESTful web service for Arabic NER is developed and deployed online\footnote{\scriptsize{ \url{https://ontology.birzeit.edu/wojood}}} as part of our language understanding resources \cite{JA19,HJ21,JAM19,JKKS21}. The web service takes a text as input and returns the output in three different formats: (\textit{i}) JSON IOB2, a JSON in which each token in the input text is returned with its corresponding tag similar to the IOB2 scheme, (\textit{ii}) JSON entities, only the recognized named entities and their positions are returned, and (\textit{iii}) XML, which is similar to the format (\textit{ii}), but the named entities are marked up using XML. Additionally, as Figure \ref{fig:demo} illustrates, a user interface is developed on top the web service for demonstration purposes, in which nested named entities are highlighted.

\section{Conclusions and Future Work}
\label{sec:conclusion}
In this paper, we have presented Wojood, a new large corpus for Arabic nested NER, which consists of 550K tokens collected from MSA and dialect text in multiple domains, which was manually annotated using a rich set of 21 entity types. The IAA evaluation of the corpus, using $Kappa$ and F1-score, illustrated a very high agreements between the annotators. We also trained a nested NER model based on multi-task learning using AraBERT, in which we achieved an F1-score of 0.884.

We plan to enrich the corpus in several directions. We are working to increase the size of the corpus by including more specialized domains and new entity types. We also plan to manually link the extracted entities with nodes in the Wikidata Knowledge Graph in order to facilitate Arabic Entity Linking research. Another natural extension to this work is relationship extraction, but this annotating the relationships first. We will also address the limitation of our model architecture so that it can predict nested entities of the same type. Last but not least, we plan on linking the entities to concepts in the Arabic ontology \cite{J21,J11} to enable richer semantic understanding of text.

\section*{Acknowledgements}
This research is piratically funded by the Palestinian Higher Council for Innovation and Excellence. We would like to thank Taymaa Hammouda and Wasim Khatib for the technical support, and Prof. Fadi Zaraket for suggestions in the early phases. The authors also acknowledge the great efforts of many students who helped in the annotation process, especially Rania Shahwan, Donia Baninemra, Shimaa Hamayel, Ayah Hamdan, Jenan Morrar, and Sondos Ilaiwi.

\section{Bibliographical References}\label{reference}

\bibliographystyle{lrec2022-bib}
\bibliography{references_lrec2022, MyReferences}
\bibliographystylelanguageresource{lrec2022-bib}
\end{document}